\documentclass[letter, 10 pt, conference]{ieeeconf}  

\IEEEoverridecommandlockouts                              

\let\labelindent\relax
\usepackage{enumitem}
\usepackage{graphics} 
\usepackage{epsfig} 
\usepackage{mathptmx} 
\usepackage{times} 
\usepackage{amsmath} 
\usepackage{amssymb}  
\usepackage{amsfonts}
\usepackage{graphicx}
\usepackage{tabularx}
\usepackage{multicol}
\usepackage{algorithm}
\usepackage{algorithmic}
\usepackage{multirow}
\usepackage{todonotes}%
\usepackage{lettrine}
\usepackage{subfigure}
\usepackage{wasysym}

\usepackage[font={scriptsize,it}]{caption} 
\usepackage[linewidth=1pt]{mdframed}

\newcommand{\x}{\mathbf{x}}

\newcommand{\z}{\mathbf{z}}

\usepackage{cite}
\usepackage[bookmarks=false]{hyperref}
\usepackage{leftidx}

\definecolor{lightlightgrey}{rgb}{0.9,0.9,0.9}
\definecolor{Red}{rgb}{1,0,0}
\definecolor{Blue}{rgb}{0,0,1}
\definecolor{Green}{rgb}{0,1,0}
\definecolor{magenta}{rgb}{1,0,.6}
\definecolor{lightblue}{rgb}{0,.5,1}
\definecolor{lightpurple}{rgb}{.6,.4,1}
\definecolor{gold}{rgb}{.6,.5,0}
\definecolor{orange}{rgb}{1,0.4,0}
\definecolor{hotpink}{rgb}{1,0,0.5}
\definecolor{newcolor2}{rgb}{.5,.3,.5}
\definecolor{newcolor}{rgb}{0,.3,1}
\definecolor{newcolor3}{rgb}{1,1,1}
\definecolor{darkgreen1}{rgb}{0, .35, 0}
\definecolor{darkgreen}{rgb}{0, .6, 0}
\definecolor{darkred}{rgb}{.75,0,0}

\xdefinecolor{olive}{cmyk}{0.64,0,0.95,0.4}
\xdefinecolor{purpleish}{cmyk}{0.75,0.75,0,0}

\makeatletter
\IEEEtriggercmd{\reset@font\normalfont\fontsize{7.9pt}{8.40pt}\selectfont}
\makeatother
\IEEEtriggeratref{1}

\xdefinecolor{red}{rgb}{1,0,0}
\mdfsetup{
  align=center,
  backgroundcolor=lightlightgrey,
  rightmargin = 5pt,
  leftmargin = 5pt,
  linecolor=red!60!black,
  linewidth=0.1pt,
  topline=false,
  rightline=false,
  ntheorem = false, 
  innerleftmargin = 0pt,
  innerrightmargin = 0pt,
  leftline=false,
  nobreak=false}

\usepackage{float}
\floatstyle{boxed}
\newfloat{math}{thp}{lop}
\floatname{math}{Equation}

\makeatother

\begin{document}

\title{Deep Neural Network-based Cooperative Visual Tracking through \\ Multiple Micro Aerial Vehicles}

\author{Eric Price$^1$, Guilherme Lawless$^1$, Heinrich H. B\"ulthoff$^2$, Michael Black$^1$ and Aamir Ahmad$^1$
\thanks{eric.price,guilherme.lawless,hhb,black,aamir.ahmad@tuebingen.mpg.de}
\thanks{$^1$Max Planck Institute for Intelligent Systems, T\"ubingen, Germany.}
\thanks{$^2$Max Planck Institute for Biological Cybernetics, T\"ubingen, Germany.}}
\maketitle
\thispagestyle{plain}
\pagestyle{plain}
\begin{abstract}


Multi-camera full-body pose capture of humans and animals in outdoor environments is a highly challenging problem. Our approach to it involves a team of cooperating micro aerial vehicles (MAVs) with on-board cameras only. The key enabling-aspect of our approach is the on-board person detection and tracking method. Recent state-of-the-art methods based on deep neural networks (DNN) are highly promising in this context. 
However, real time DNNs are severely constrained in input data dimensions, in contrast to available camera resolutions. Therefore, DNNs often fail at objects with small scale or far away from the camera, which are typical characteristics of a scenario with aerial robots. Thus, the core problem addressed in this paper is how to achieve on-board, real-time, continuous and accurate vision-based detections using DNNs for visual person tracking through MAVs. Our solution leverages cooperation among multiple MAVs. First, each MAV fuses its own detections with those obtained by other MAVs to perform cooperative visual tracking. This allows for predicting future poses of the tracked person, which are used to selectively process only the relevant regions of future images, even at high resolutions. Consequently, using our DNN-based detector we are able to continuously track even distant humans with high accuracy and speed. We demonstrate the efficiency of our approach through real robot experiments involving two aerial robots tracking a person, while maintaining an active perception-driven formation. Our solution runs fully on-board our MAV's CPU and GPU, with no remote processing. ROS-based source code is provided for the benefit of the community.

\end{abstract}

\section{Introduction}

Human/animal pose tracking and full body pose estimation and reconstruction in outdoor, unstructured environments is a highly relevant and challenging problem. Its wide range of applications includes managing large public gatherings, search and rescue \cite{uav_victim_detection,uav_victim_search}, coordinating outdoor sports events \cite{MULTIDRONE} and facilitating animal conservation efforts in the wild \cite{uav_wildlife}. In indoor settings, similar applications usually make use of body-mounted sensors, artificial markers and static cameras. While such markers might still be usable in outdoor scenarios, dynamic ambient lighting conditions and the impossibility of having static/fixed cameras make the overall problem difficult. On the other hand, body-mounted sensors are not suitable for some kinds of subjects (e.g., animals in the wild or large crowds of people). Therefore, our approach to the aforementioned problem involves a team of micro aerial vehicles (MAVs) tracking subjects by using only on-board monocular cameras and computational units, without any subject-fixed sensor or marker. Among the several challenges involved in developing such a system, multirobot cooperative detection and tracking (MCDT) of a subject's 3D position is one of the most important. It is also the sole focus of this paper.

\smallskip

\begin{figure}
 \includegraphics[width=\columnwidth, height=4.5cm]{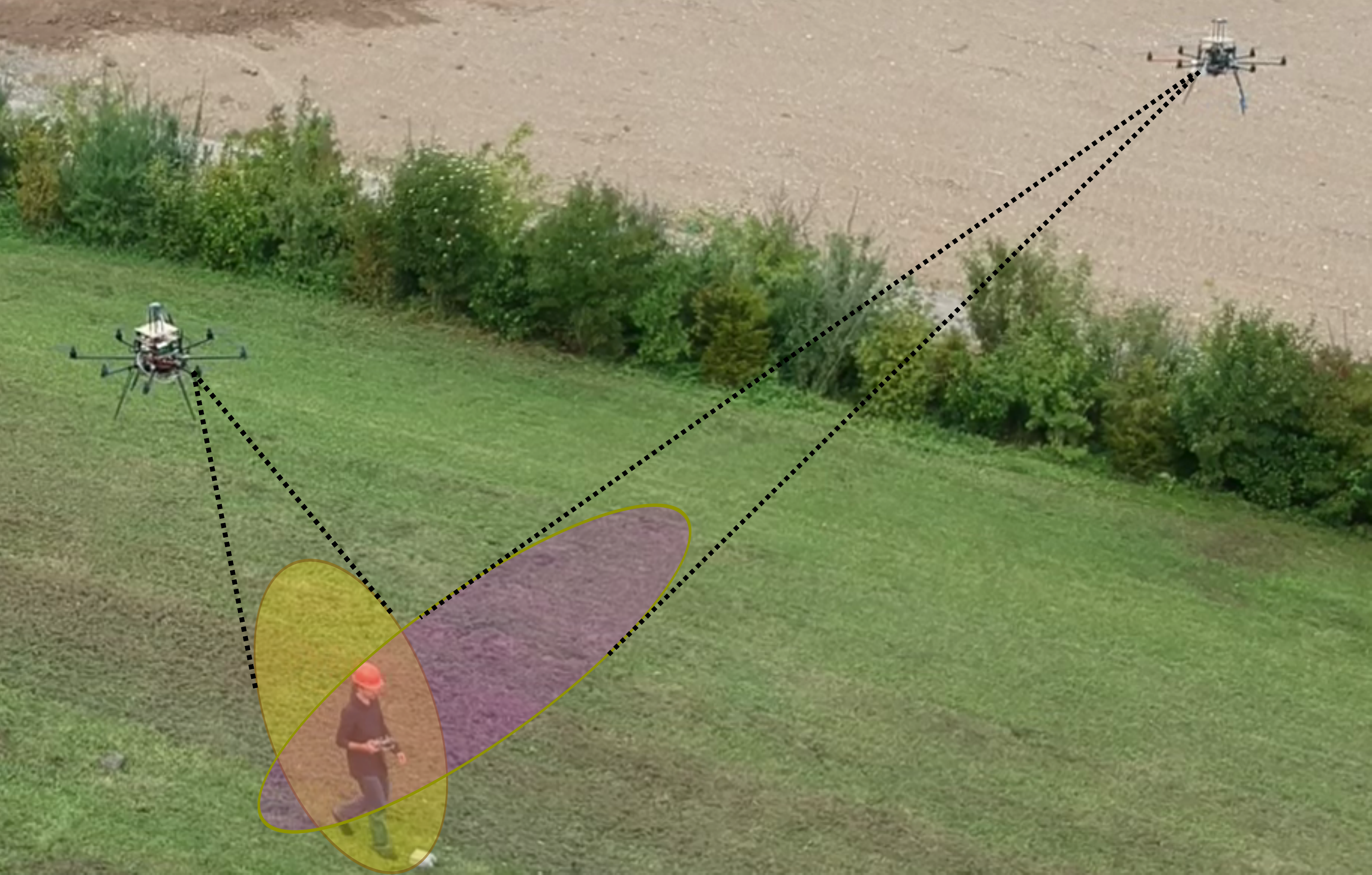}
 \caption{Two octocopters cooperatively detecting and tracking a person while maintaining a perception-driven formation.}
\end{figure}

The key component of our MCDT solution is the person detection method suitable for outdoor environments and marker/sensor-free subjects. Deep convolutional neural network-based (DNN) person detection methods are, unarguably, the state-of-the-art. In recent past, DNNs have consistently shown to outperform traditional techniques for object detection \cite{schmidhuber2015deep}. Consequently, they have gained immense popularity in various robotics-related fields, such as, autonomous driving \cite{falcini2017deep} and detection and identification of pedestrians \cite{hosang2015taking} and vehicles \cite{sun2006road}. However, there are only few works that exploit the power of DNNs for visual object detection on board MAVs.

\smallskip

The main limiting factor in using DNNs on MAVs is the computational requirements of these networks. The majority are not real-time capable. The fastest (open-source) DNN-based detector \cite{liu2016ssd} achieves approximately $60$ frames per second (fps) on a $300\times 300$ pixel square area. However, this requires the computing power of a dedicated high end GPU, which is usually bulky and impractical to install on-board a MAV. Moreover, the higher power consumption of such a GPU often requires a bulkier power supply, further aggravating this issue. One alternative is transmitting raw images to a ground-station in order to offload their processing. However, this approach is also infeasible due to communication bandwith constraints when multiple MAVs are used. Using the most advanced\footnote{Available at the time of performing the work presented in this paper.} light weight GPU (Jetson TX1 \cite{otterness17evaluation}), Wei Liu et al.'s single shot Multibox detector (SSD Multibox) achieves $\sim 4$ fps \cite{liu2016ssd} in sequential processing. This hardware is suitable for airborne use, but the speed is just at the lower boundary of real time applicability.

\smallskip

 On the other hand, the $4$ fps performance of SSD is only with low resolution images. Real-time on-board processing of high resolution images is still infeasible. Thus, unfortunately, it cannot exploit the richness of very high-resolution images provided by most modern cameras. A na\"ive approach to use high resolution images will be to perform an uninformed and non-selective processing of a down-sampled, low resolution input image. This will always be suboptimal due to the reduced information content. Also, a low resolution image could have a wide field of view (FOV), but often lacks in detail regarding a distant subject, thus limiting detection capability and accuracy. Alternatively, an un-informed region of interest (ROI) (e.g., around the center of the original image) greatly limits the chance of the tracked person being in the FOV. Therefore, DNNs would often fail at objects with small scale or far away from the camera, which are typical characteristics of a scenario with aerial robots.

\smallskip

To achieve real-time, continuous and accurate vision-based detections using DNNs for MCDT, our solution leverages the mutual world knowledge which is jointly acquired by our multirobot system during cooperative person tracking. However, this does not imply sharing full resolution raw images at high frame rates between MAVs. In a multirobot environment, the required communication bandwidth would increase at least linearly with each additional robot, rendering such data sharing impractical. Instead, we only share compactly-representable detection measurements which have low bandwidth requirements. Using these, we perform cooperative person tracking and use the predicted poses of the tracked person to feed the DNN-based detector a well-informed ROI for future images. Consequently, each MAV knows where to look next and can actively select the relevant ROI that supplies the highest information content. Thus, our method not only minimizes the information loss incurred by downsampling the high resolution image, but also maximizes the chance of the person being completely in the FOV. Fusion of detection measurements within our MCDT method takes into account the associated person detection uncertainties as well as the self-localization uncertainties of all MAVs.

\smallskip

In this work we use SSD Multibox detector \cite{liu2016ssd} due to its versatility in detecting a wide range of object classes, availability of annotated training data, speed and accuracy. However, it is important to note that our approach is not tailored to this specific detector. We treat it as a black box over a defined interface. Any visual detector can be used, as long as it can operate on arbitrarily scaled subsets of the input space. 

\bigskip


The core contributions of this paper are as follows.
\begin{itemize}
 \item A real-time, continuous and accurate DNN-based multi-robot cooperative detection and tracking method, suitable for MAVs with only on-board camera and computation. This method
 \begin{itemize}
  \item allows increased sensitivity to small and far away targets compared to a fixed wide-angle low-resolution detector,
  \item allows increased field of view compared to a fixed narrow-angle low-resolution detector,
  \item performs significantly better than exhaustive searches over the entire high resolution image, such as using tiled or sliding window approaches, and
  \item keeps computational complexity independent of the camera image resolution.
 \end{itemize}
 \item A method for characterizing the DNN-based detection measurement noise.
 \item Fully open source ROS-based implementation of our method.  
\end{itemize}

\smallskip

We evaluate the performance of our MCDT approach through active cooperative perception experiments. A team of two 8-rotor MAVs actively detects and tracks a person in an outdoor environment. We demonstrate that using our MCDT method, the MAVs are able to consistently track and follow a person, while i) jointly maintaining a high quality 3D position estimate of the person, ii) remaining oriented towards the person, and iii) satisfying other formation constraints.
The formation controller is based on one of our previous works \cite{aamir_pcmmc_ras_journal}. It involves the combination of a model predicted controller (MPC) and a potential field algorithm. In the next section we situate our work within the state-of-the-art. This is followed by a description of our system design, theoretical details of our proposed MCDT approach and characterization of detection measurement errors in Sec.~\ref{sec:theory}. Section~\ref{sec:exps} presents our experiments and results, followed by conclusions in Sec.~\ref{sec:conc}.

\section{State-of-the-Art}
\label{sec:sota} 

Motivated by several applications in graphics and computer vision, e.g., virtual reality, sports analysis, pose monitoring for health purposes and crowd management, full body pose and motion estimation has attracted widespread research attention \cite{pose_estimation_review}. Most state-of-the-art research in this field is focused on indoor scenarios, e.g., \cite{indoor_reconstruction_bogo}. However, recent works have started to address the challenges of performing full body motion capture in unstructured everyday environments \cite{motion_cap_everyday_env} and even in unknown outdoor scenarios \cite{SIP_gerard}. Authors in \cite{SIP_gerard} use body-fixed sensors (IMUs) to estimate the full body pose. Although they achieve a high degree of accuracy, the setup itself may be cumbersome or even infeasible for certain subjects (e.g., animals). To avoid this, as well as any body-fixed markers, the alternative approach is to perform motion capture using only multiple cameras.
In \cite{multiview_camera_mocap}, authors presented such a method to extract the skeletal motion of multiple closely-interacting people by using multiple cameras, without using body-fixed markers or sensors. However, their solution assumes a scenario where cameras can be fixed to the environment and the ambient light is controlled. In unknown outdoor environments, these assumptions do not hold. Hence, our approach to motion capture in outdoor scenarios involves multiple cameras on board a fleet of MAVs. In this paper, we address the key issue of multirobot cooperative detection and tracking (MCDT) involved in developing such an outdoor motion capture system. Through a cooperative perception-driven formation, we not only ensure that the tracked subject is almost always in the FOV of all MAVs, but also prevent the MAVs from colliding with each other. 

Multirobot cooperative tracking has been researched extensively over the last years \cite{recent_cooperative_target_tracking_paper_13,aahmad_tro,ctt_multiple_robs_IEEE_12}. While the focus of most of these methods is to improve the tracked target's pose estimate by fusing information obtained from teammate robots, some recent methods, e.g., \cite{aahmad_tro}, simultaneously improve the localization estimates of the poorly localized robots in addition to the tracked target's pose, within a unified framework. However, hardly any cooperative target tracking method directly facilitates detections through cooperation among the robots. In this paper we do so by sharing independently obtained measurements among the robots, regarding a mutually observed object in the world. Using these shared measurements, our MCDT method at each MAV allows its detector to focus only on the relevant and most informative ROIs for future detections.

Cooperative tracking of targets using multiple MAVs has also attracted attention in the recent past \cite{hausman2016cooperative,swarm_coverage,fixed_wings_target_tracking_zhang_16}. While some address the problem of on-board visual detection and tracking using MAVs \cite{fu2016onboard}, they still rely on hand-crafted visual features and traditional detection approaches. Moreover, cooperation among multiple MAVs using on-board vision-based detections is not well addressed. In this paper we address both of these issues by using a DNN-based person detector that runs on board the MAVs and sharing the detection measurements among the MAVs to perform MCDT.

Deep CNN-based detectors currently require the power of GPUs. For MAVs, light-weight GPUs or embedded solutions are critical requirements. In \cite{rallapalli2016very}, Rallapalli et al.\ present an overview of the feasibility of deep neural network-based detectors for embedded and mobile devices. Also, several recent works now concern airborne applications of GPU-accelerated neural networks for computer vision tasks. However, they mostly evaluate their performance in offline scenarios, without a flight capable implementation \cite{de2016towards,zhang2016deep}. On the other hand, there are some networks suitable for real time detection and localization of arbitrary objects in arbitrary poses and backgrounds. These include networks, such as, YOLO \cite{redmon2016you} or Faster R-CNN \cite{ren2015faster}, which are both outperformed in speed and detection accuracy by SSD Multibox approach \cite{liu2016ssd}. Hence, in our work we use the latter. Furthermore, we ensure its real-time operation using a camera of any given resolution. We achieve this through our cooperative approach involving active selection of the most informative ROI, a method that is novel to the best of our knowledge.

\section{System Overview and the Proposed Approach}
\label{sec:theory} 

\subsection{System Overview}
\label{sec:sysoverview}
\begin{figure}[t!]
\includegraphics[width=\columnwidth]{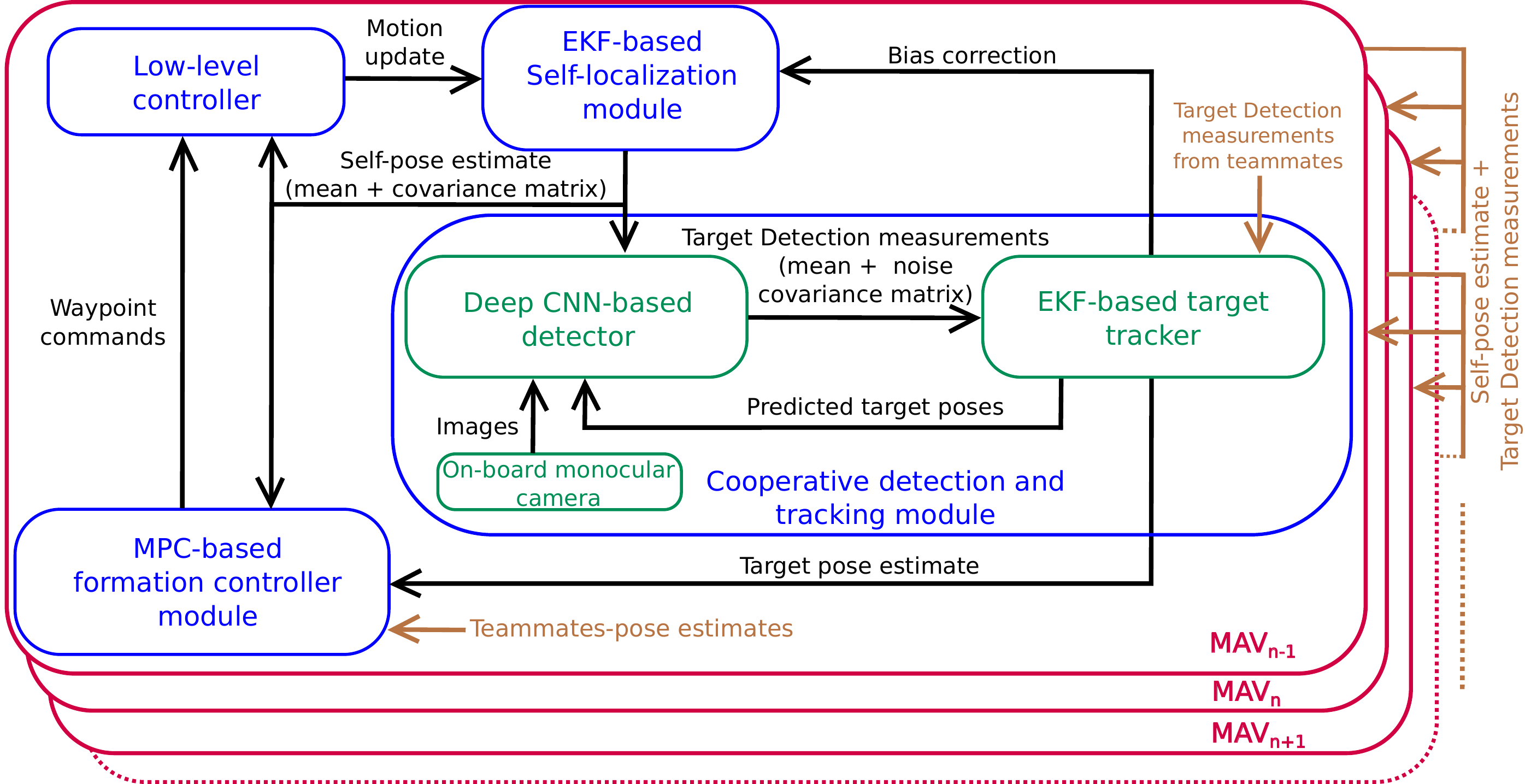}
\caption{Overall architecture of our multi-MAV system, designed to cooperatively detect and track a person while maintaining a perception-driven formation and satisfying certain formation constraints. The focus of the work presented in this paper and the novelties correspond to the green blocks in this figure.}
\label{fig:system}
\end{figure}

Our multi-MAV system does not consist of a central computational unit. Each of our MAVs is equipped with an on-board CPU and GPU to perform all computations. Although our architecture also does not depend on a centralized communication network, the field implementation is done through a central wifi access point. Each MAV runs its own instance of the following software modules. A low-level flight-controller module, a self-localization module, a cooperative detection and tracking (CDT) module and an MPC-based formation controller module. Figure~\ref{fig:system} details the flow of data among these modules. The data shared between any two MAVs consists of their self-pose estimates and the detection measurements of the tracked person. All the aforementioned software modules run on board in realtime. In the following sections, after introducing notations, we focus on the detailed description of our CDT module. Therein we describe how our proposed MCDT approach enables the MAV formation to track seemingly-small and far-away persons with high accuracy without losing them from any MAV's FOV during tracking.

\setlength{\belowcaptionskip}{0pt}
\setlength{\belowcaptionskip}{-20pt}

\subsection{Preliminaries}
\label{sec:preliminaries}

Let there be $N$ MAVs $R_1,..., R_N$ tracking a person $P$. Let the 6D pose (3D position and 3 orientation angles in body-frame coordinates) of the $n_{\text{th}}$ MAV in the world frame at time $t$ be given by $^{W}\x_t^{R_n} \in \mathbb{R}^6$, obtained using a self-localization system. The uncertainty covariance matrix associated to the MAV pose is given as $^{W}\boldsymbol{\Sigma}_t^{R_n} \in \mathbb{R}^{6\times6}$.
Each MAV $n$ has an on-board, monocular, perspective camera $C_n$, rigidly attached to it.

Let the position and velocity of the tracked person in the world frame at time $t$ be given by $^{W}\x_t^{P}\in \mathbb{R}^6$. We assume that the person is represented by the position and velocity of its centroid in the 3D Euclidean space. The uncertainty covariance matrix associated to it is given by $^{W}\boldsymbol{\Sigma}_t^{P} \in \mathbb{R}^{6\times6}$.

\subsection{DNN-based Cooperative Detection and Tracking}

\begin{algorithm}[t!]
\caption{Cooperative Detector and Tracker (CDT) on MAV $R_k$ with inputs $\lbrace^{W}\x_{t-1}^{P}, ^{W}\boldsymbol{\Sigma}_{t-1}^{P}, \mathbf{ROI}_{t-1}^{R_k}, \mathbf{I}_{t}^{R_k}\ \rbrace$ } 
\begin{algorithmic}[1]
\label{Alg:cdt}
\STATE $\lbrace \leftidx{^W}\z_t^{P,R_k}, {}^{W}\boldsymbol{Q}_t^{P,R_k}\rbrace \leftarrow $ DNN person detector $(\mathbf{I}_{t}^{R_k},\mathbf{ROI}_{t-1}^{R_k})$.
\STATE $\mathbf{Transmit}~~$  $\lbrace ^{W}\x_t^{R_k}, ~ ^{W}\boldsymbol{\Sigma}_t^{R_k}, ~ ^{W}\z_t^{P,R_k} $ and $  {}^{W}\boldsymbol{Q}_t^{P,R_k}  \rbrace$  to all MAVs $\mathrm{R}_n;~ n=1,...,N;~n\neq k$ \\ $~$
\STATE $\mathbf{Receive}~~$  $\lbrace ^{W}\x_t^{R_n}, ~ ^{W}\boldsymbol{\Sigma}_t^{R_n}, ~ ^{W}\z_t^{P,R_n} $ and $  ^{W}\boldsymbol{Q}_t^{P,R_n}  \rbrace$ such that $\mathrm{R}_n;~ n=1,...,N;~n\neq k$ \\ $~$
\STATE $\lbrace^{W}\bar{\x}_t^{P}, ^{W}\bar{\boldsymbol{\Sigma}}_t^{P}\rbrace \leftarrow $ EKF Prediction $\lbrace^{W}\x_{t-1}^{P}, ^{W}\boldsymbol{\Sigma}_{t-1}^{P}\rbrace$ using exponentially decelerating state transition model.
\FOR{$n = 1$ to $N$}
   \STATE \hspace*{-12pt}  $\lbrace^{W}\x_t^{P}, ^{W}\boldsymbol{\Sigma}_t^{P}\rbrace \leftarrow $ EKF Update $\lbrace^{W}\bar{\x}_{t}^{P}, \leftidx{^W}{\boldsymbol{\bar{\Sigma}}}_t^{P}, \leftidx{^{W}}\z_t^{P,R_n},  \leftidx{^{W}}{\boldsymbol{Q}}_t^{P,R_n}  \rbrace$
\ENDFOR
\vspace*{3pt}
\STATE $\lbrace^{W}\hat{\x}_{t+1}^{P}, ^{W}{\boldsymbol{\hat{\Sigma}}}_{t+1}^{P}\rbrace \leftarrow $ Predict for next ROI $\lbrace^{W}\x_t^{P}, ^{W}\boldsymbol{\Sigma}_t^{P}\rbrace$
\vspace*{3pt}
\STATE $\mathbf{ROI}_{t}^{R_k} \leftarrow$ Calculate next ROI$\lbrace^{W}\hat{\x}_{t+1}^{P}, ^{W}{\boldsymbol{\hat{\Sigma}}}_{t+1}^{P}\rbrace$
\vspace*{3pt}
\STATE Update self-pose bias$\lbrace\leftidx{^{W}}\z_t^{P,R_k} $, $  \leftidx{^{W}}{\boldsymbol{Q}}_t^{P,R_k},\leftidx{^{W}}\x_t^{P}, \leftidx{^{W}}{\boldsymbol{\Sigma}}_t^{P}\rbrace$
\vspace*{3pt}
\RETURN $\lbrace^{W}\x_t^{P}, ^{W}\boldsymbol{\Sigma}_t^{P}, \mathbf{ROI}_{t}^{R_k}\rbrace$
\end{algorithmic}
\end{algorithm} 

\setlength{\textfloatsep}{5pt}

Algorithm~\ref{Alg:cdt}, which is a recursive loop, outlines our CDT approach. Each MAV runs an instance of this algorithm in real time. The algorithm is based on an EKF where the inputs are the person's tracked 3D position estimate $^{W}\x_{t-1}^{P}$ at the previous time step $t-1$, the covariance matrix $^{W}\boldsymbol{\Sigma}_{t-1}^{P}$ associated to that estimate and $\mathbf{ROI}_{t-1}^{R_k}$, also computed at $t-1$. The other input is the image $\mathbf{I}_{t}^{R_k}$ at $t$. Steps~1--11 correspond to the iteration of the algorithm at $t$.

Step~1 of Alg.~\ref{Alg:cdt} performs the person detection using a DNN-based detector on a ROI $\mathbf{ROI}_{t-1}^{R_k}$ provided to it from the previous time step $t-1$ and the image $\mathbf{I}_{t}^{R_k}$ at the current time step $t$. Using raw detection measurements on the image (explained further), the detection measurement in the world frame is computed as a mean $\leftidx{^W}\z^{P,R_k}$ and a noise covariance matrix $\leftidx{^W}{\boldsymbol{Q}}_t^{P,R_k}$. The detection is performed on the latest available image and uses only the GPU. If this resource is busy processing a previous images, the detection (and therefore the EKF update, Step 6) is skipped. 

Raw detection measurements on the image consist of a set of rectangular bounding boxes with associated confidence scores and a noise covariance matrix. To obtain this noise, we performed a characterization of the DNN-based detector, explained in Sec.~\ref{subsec:dnnNoise}. The raw detection measurements are transformed first to the camera coordinates and finally to the world coordinates. This transformation incorporates the noise covariances in the raw detection measurements and the MAV's self-pose's uncertainty covariance. Note that the raw measurements are in 2D image plane, where as the final detection measurements are computed in the world frame. For this, we make a further assumption on the height of the person being tracked. We assume that the person's height follows a distribution $H \sim \mathcal{N}(\mu_H,\,\sigma^{2}_H)$. While we assume that the person is standing or walking upright in the world frame, the model can be adapted to consider a varying pose (e.g., sitting down), for instance, by increasing $\sigma^{2}_H$. Finally the overall transformation of detections from image frame to world frame measurements also takes $\sigma^{2}_H$ into account. For mathematical details regarding the transformation that include propagating noise/uncertainty covariances, we refer \cite{aamir_stereo_fusion}.

In Steps~2--3 of Alg.~\ref{Alg:cdt} we transmit and receive data among the robots. This includes self-pose estimates (used for inter-robot collision avoidance) and the detection measurements, both in world frame.
Step~4 performs the prediction step of the EKF. 
Here we use an exponentially decelerating state transition model. When the algorithm continuously receives measurements allowing continuous update steps, predictions behave similar to a constant velocity model. However, when the detections measurements stop arriving, the exponential decrease in velocity allows the tracked person's estimate to become stationary. This is an important property of our CDT approach, as the ROI is calculated from the tracked person's position estimate (Step~8). In the case of having no detection measurements, the uncertainty in the person's position estimate would continuously grow, resulting in a larger ROI. This is further clarified in the explanation of Steps~8--9.

In Steps~5--7 we fuse measurements from all MAVs (including self-measurements). As these measurements are in the world frame, fusion is done by simply performing an EKF update for each measurement.

In Steps~8--9 of Alg.~\ref{Alg:cdt} lies the key novelty of our approach. We actively selects a ROI ensuring that future detections are performed on the most informative part of the image, i.e., where the person is, while keeping the computational complexity independent of camera image resolution. As the computational complexity of DNN-based detectors grows very fast with the image resolution, using our approach we are still able to use a DNN-based method in real-time and with high detection accuracy. The ROI is calculated as follows. 
First (Step 8), using a prediction model similar to the EKF prediction and the estimates at the current time step ($\leftidx{^{W}}\x_t^{P}$ and $\leftidx{^{W}}{\boldsymbol{\Sigma}}_t^{P}$), we predict the state of the person in the next time step $t+1$ as $\lbrace^{W}\hat{\x}_{t+1}^{P}, ^{W}{\boldsymbol{\hat{\Sigma}}}_{t+1}^{P}\rbrace$.
Then, in Step 9, using the predicted 3D position of the person (only the position components of $\lbrace^{W}\hat{\x}_{t+1}^{P}, ^{W}{\boldsymbol{\hat{\Sigma}}}_{t+1}^{P}\rbrace$), we calculate the position and associated uncertainty of the person's head $\lbrace\leftidx{^{W}}{\hat\x}_{t+1}^{P_h}$, $\leftidx{^{W}}{\boldsymbol{\hat\Sigma}}_{t+1}^{P_h}\rbrace$ and feet $\lbrace\leftidx{^{W}}{\hat\x}_{t+1}^{P_f}$, $\leftidx{^{W}}{\boldsymbol{\hat\Sigma}}_{t+1}^{P_f}\rbrace$. For this we again assume the height distribution model for a person, as introduced previously, and that the person is in an upright position. $\leftidx{^{W}}{\hat\x}_{t+1}^{P}$, $\leftidx{^{W}}{\hat\x}_{t+1}^{P_h}$ and $\leftidx{^{W}}{\hat\x}_{t+1}^{P_f}$ are back-projected onto the image frame along with the uncertainties and are denoted as $\lbrace\leftidx{^{I}}{\hat\x}_{t+1}^{P}$, $\leftidx{^{I}}{\boldsymbol{\hat\Sigma}}_{t+1}^{P}\rbrace$ (person's center), $\lbrace\leftidx{^{I}}{\hat\x}_{t+1}^{P_h}$, $\leftidx{^{I}}{\boldsymbol{\hat\Sigma}}_{t+1}^{P_h}\rbrace$ (head) and $\lbrace\leftidx{^{I}}{\hat\x}_{t+1}^{P_f}$, $\leftidx{^{I}}{\boldsymbol{\hat\Sigma}}_{t+1}^{P_f}\rbrace$ (feet). The center-top pixel of the ROI is now given by
\begin{equation}
 \mathbf{ROI}_{t}^{R_k}\{\textbf{center, top}\} = (\leftidx{^{I}}{\hat{x}}_{t+1}^{P}, \leftidx{^{I}}{\hat{y}}_{t+1}^{P_h} + 3 ~\leftidx{^{I}}{{\boldsymbol{{\hat{\sigma}}_y}}}_{t+1}^{P_h})
\end{equation}
and the the center-bottom pixel given by
\begin{equation}
 \mathbf{ROI}_{t}^{R_k}\{\textbf{center, bottom}\} = (\leftidx{^{I}}{\hat{x}}_{t+1}^{P}, \leftidx{^{I}}{\hat{y}}_{t+1}^{P_f} - 3 ~\leftidx{^{I}}{{\boldsymbol{{\hat{\sigma}}_y}}}_{t+1}^{P_f}),
\end{equation}where $\leftidx{^{I}}{\hat{x}}_{t+1}^{P}$ is the x-axis component of $\leftidx{^{I}}{\hat\x}_{t+1}^{P}$.
$\leftidx{^{I}}{\hat{y}}_{t+1}^{P_h}$ and $\leftidx{^{I}}{\hat{y}}_{t+1}^{P_f}$ are the y-axis components of $\leftidx{^{I}}{\hat\x}_{t+1}^{P_h}$ and $\leftidx{^{I}}{\hat\x}_{t+1}^{P_f}$, respectively. $\leftidx{^{I}}{{\boldsymbol{{\hat{\sigma}}_y}}}_{t+1}^{P_h}$ and $\leftidx{^{I}}{{\boldsymbol{{\hat{\sigma}}_y}}}_{t+1}^{P_f}$ are the standard deviations in the y-axis computed directly from  $\leftidx{^{I}}{\boldsymbol{\hat\Sigma}}_{t+1}^{P_h}$ and $\leftidx{^{I}}{\boldsymbol{\hat\Sigma}}_{t+1}^{P_f}$, respectively. Lastly, the left and right borders of the ROI are calculated to match a desired aspect ratio. We chose the aspect ratio $(4:3)$ that corresponds to the majority of the training images for optimal detection performance.

Step~10 of Alg.~\ref{Alg:cdt} performs the bias update of the MAV self-pose. Self-pose estimates obtained using GPS and IMU sensors (as is the case of our self-localization system) often have a time-varying bias \cite{whitacre2009cooperative}. The self-pose biases of each MAV cause mismatches when fusing detection measurements in the world frame, which are detrimental to our MCDT approach because i) the calculated ROI for person detection will be biased and will often not include the person, and ii) the person's 3D position estimate will be a result of fusing biased measurements. To truly benefit from our MCDT approach, we track and compensate the bias in each MAV's localization system (see \ref{sec:sysoverview}). We track these biases by using an approach based on \cite{whitacre2009cooperative}. The difference between a MAV's own detection measurements and the tracked estimate (after fusion) is used to update the bias estimate. Our approach is a simplified form of \cite{whitacre2009cooperative} as we only track position biases, which is sufficient if the bias in orientation is negligible.

\subsection{Noise Quantification of a DNN-based Object Detector}
\label{subsec:dnnNoise}

Our MCDT approach depends on a realistic noise model of the person detector. To this end, we perform its noise quantification. As this is dependent on the specific detector being used, we describe the procedure followed for the pre-trained SSD Multibox detector\cite{liu2016ssd}, used in our work.

SSD Multibox can detect the presence and location of a set of object classes in an image, in which people are included. The output of the detector is, for each detection, a bounding box, a class label and a confidence score. The input image size is defined at training time. We used a pre-trained\footnote{SSD300 trained on the PASCAL VOC 2007+12 dataset. Available: \url{https://github.com/weiliu89/caffe/tree/ssd}} $300\times300 px$ network in this work.

\begin{figure}
	\begin{center}
		\includegraphics[width=0.7\columnwidth]{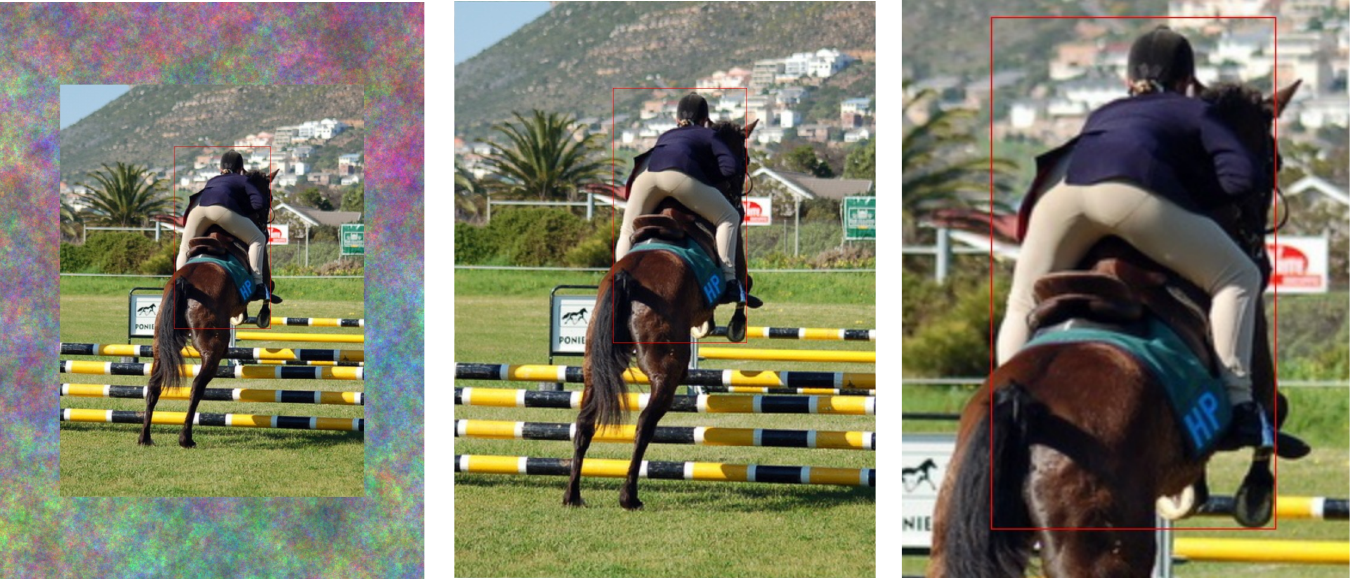}
		\caption{Example of our extended test set created from the PASCAL VOC 2007 dataset. From the left: downscaled, original and upscaled images.}
		\label{fig:testset}
	\end{center}
\end{figure}

We quantify the detection noise with respect to (w.r.t.) the size of the detections, i.e., smaller and distant, or larger and closer to the camera. To this end, we created an extended test set from the PASCAL VOC 2007 dataset with varying levels of downscaling and upscaling of the detected person. Figure \ref{fig:testset} details one such example. We selected images with a single, non-truncated person to avoid the detection association problem.

\setlength{\belowcaptionskip}{-5pt}

\begin{figure}
	\begin{center}
		\includegraphics[width=1\columnwidth]{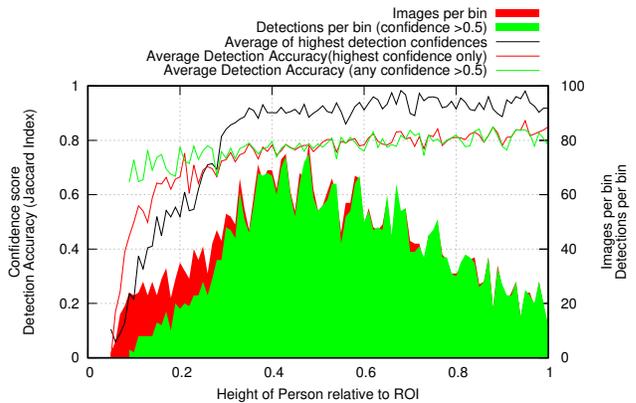}
		\caption{Detection accuracy w.r.t.\ the relative person height in the ROI. Images are divided into bins according to the height.}
		\label{fig:detectionacc}
	\end{center}
\end{figure}

Statistical analysis using the person detector in our extended test set was performed. As the test set has images of different sizes, we calculate all measures relative to their image size. Figure \ref{fig:detectionacc} shows the detection accuracy, using the Jaccard index, relative to the person height.
It is evident that even though the detection accuracy for a given minimum confidence threshold is nearly constant w.r.t.\ the analyzed ROI, the absolute error decreases with a smaller ROI. 
The chance of successful detection falls significantly for relative sizes below $30\%$ of the ROI and goes down to zero at $10\%$, thus forming an upper boundary for desired ROI size.
This analysis clearly justifies the necessity of our MCDT approach with active selection of ROIs.

\setlength{\belowcaptionskip}{-20pt}

\begin{figure}
	\begin{center}
		\includegraphics[width=1\columnwidth]{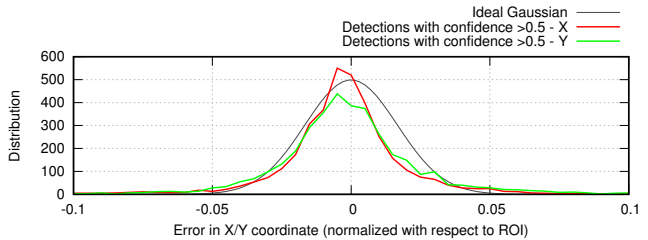}
		\caption{Error distribution of SSD Multibox detections in X and Y coordinate.}
		\label{fig:heighterrors}
	\end{center}
\end{figure}

Further analysis is presented Figure \ref{fig:heighterrors}, where we show the relative error in the detected person's position over all test images. The error is shown to be well described by a Gaussian distribution. We performed similar analysis on the errors of each detection for the top, bottom, left and right-most points of the detection bounding box at different relative sizes. We found all noises to be similarly well described by Gaussian distributions, without significant correlations between them. Thus, the person detection noise can be well approximated by a constant variance model. (see Table \ref{tab:detector-variances} for exact values).

\setlength{\belowcaptionskip}{-5pt}

\begin{table}[th!]
	\noindent \centering{}%
	\begin{tabular}{|c|c|c|c|}
		\hline 
		$\sigma^2_{top}$ & $\sigma^2_{bottom}$ & $\sigma^2_{left}$ & $\sigma^2_{right}$ \\
		\hline 
		$0.0014$ & $0.0045$ & $0.0039$ & $0.0035$ \\
		\hline 
	\end{tabular}\caption{\label{tab:detector-variances}SSD Multibox detection's noise variances for each side of the detection bounding boxes, relative to ROIs.}
\end{table}

\section{Experiments and Results}
\label{sec:exps} 

\setlength{\belowcaptionskip}{0pt}

\subsection{\label{sub:Hardware}Hardware and Software Setup}

To evaluate our approach, we conducted only real robot experiments on a team of two 8-rotor MAVs tracking a person. Each MAV is equipped with an HD camera, computer and NVIDIA Jetson TX1 embedded GPU. GPS, IMU and magnetic compass are used for self pose estimates and flight control. They use a ROS \cite{quigley2009ros} Multi-Master setup \cite{juan2015multi} over Wifi. Each MAV continuously reads images from the camera at $40Hz$. The GPU runs SSD-Multibox \cite{liu2016ssd}. As described in Sec.~\ref{sec:theory}, ROIs of images down-sampled to $300\times300$ are sent to the GPU. The detection frame rate achieved is $3.89Hz$. Detections of persons are projected into 3D, shared between the MAVs and fused using a Kalman Filter on each MAV. MAV self-pose estimates are bias corrected based on mismatches between perceived and fused 3D position of the person within the Kalman Filter. The MAVs fly in formation using an MPC based controller \cite{aamir_pcmmc_ras_journal}. The formation constraints include maintaining i) a preset distance to the tracked person's position estimate ($D_{\textit{per}}$), ii) a preset altitude above the ground plane ($H_{\textit{mav}}$), and iii) orientation towards the tracked person. For inter-robot collision avoidance we use an additional layer of artificial potential-field based avoidance algorithm on top of MPC. State estimates and IMU sensor data are updated at $100Hz$. The achieved video frame rate is $40fps$ at a resolution of $2040\times1086$ RGB. Note that during all experiments the tracked person is wearing an orange helmet only for safety. The helmet is not used for any detection. On the contrary, the DNN-based detector, which we use, was not optimized for persons with helmets.


\subsection{\label{sub:Life-Data}Online Dataset}

Using our proposed approach for MCDT and on-board real-time computation during the flights, both MAVs perform estimation of i) their 6D self-pose, and ii) 3D position of the tracked person. We treat this as our primary dataset and refer to it as the `online dataset' for analysis further in this section.

\subsection{Offline Datasets}
During each experiment, we also record raw sensor and camera data. Using these, we then generate additional \mbox{`offline datasets'}, by disabling selected aspects of our MCDT approach. This allows us to evaluate the contribution of each component of our approach to the overall performance. The following offline datasets are generated.

\begin{itemize}
 \item \textbf{No actively-selected ROI (AS-ROI)} - Here we turn off the actively selection of relevant ROI. Instead, the full image frame is down-sampled to $300\times300$ and sent to the GPU.
 \item \textbf{No MAV self-pose bias correction (SPBC)} - Here we turn off the MAV self-pose bias correction. 
 \item \textbf{Single MAV runs} - Here we perform person tracking using the data from only a single MAV. However, the active selection of ROI is still applied in the single MAV scenario.
\end{itemize}

\setlength{\belowcaptionskip}{-5pt}

\subsection{\label{sub:Static-Target-Experiment}Stationary Person Experiment}

\begin{figure*}[t]
\noindent \begin{centering}
\includegraphics[width=1\textwidth]{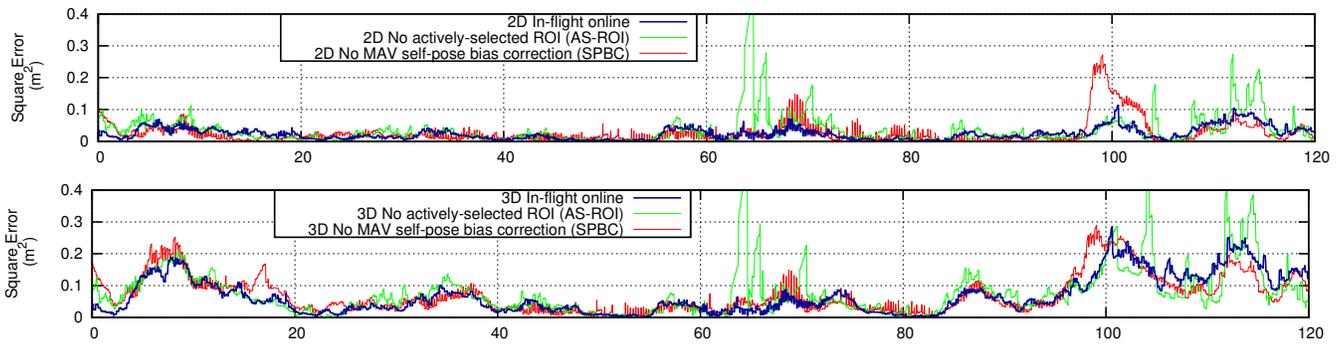}
\par\end{centering}

\noindent \centering{}\includegraphics[width=1\textwidth]{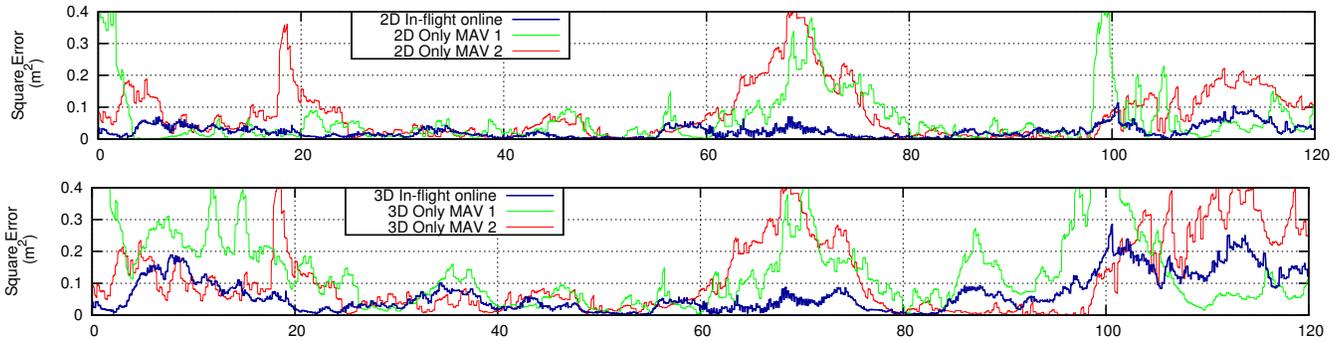}\caption{\label{fig:Position-Error-over}X Axis is time in seconds. Y axis represents 2D (only on the ground plane) or 3D position errors over time in the stationary person experiment.}
\end{figure*}
\begin{figure*}[t]
\noindent \begin{centering}
\includegraphics[width=1\textwidth]{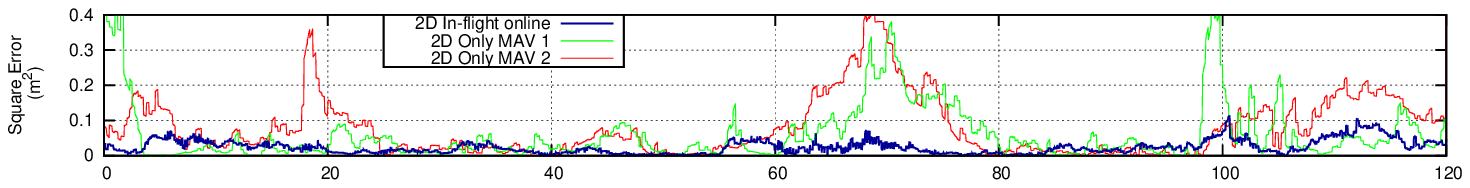}
\par\end{centering}

\noindent \centering{}\includegraphics[width=1\textwidth]{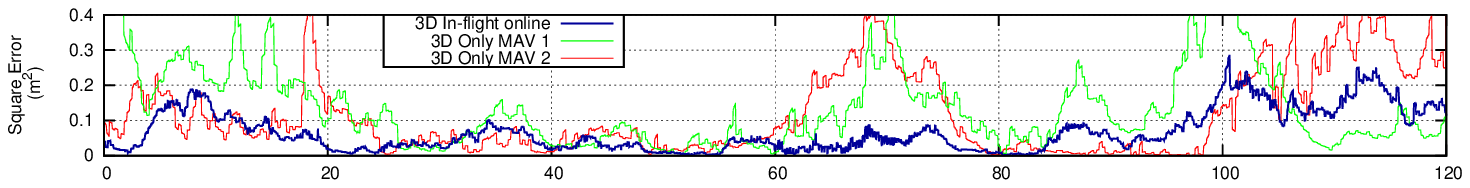}\caption{\label{fig:Position-Error-over-1}X Axis is time in seconds. Y axis represents 2D (only on the ground plane) or 3D position errors over time in the stationary person experiment.}
\end{figure*}
The formation constraints for the 2 MAVs are $D_{\textit{per}}$ = $8m$ and $H_{\textit{mav}}$ = $8m$. The person remains stationary for $120s$. We do not
have absolute ground truth measurements of the person's 3D position.
Since the person remains stationary, we instead calculate a reference
point as the mean of all 3D position estimates for each dataset (online and offline). The error
in any 3D position estimate is then calculated as its distance to this reference
point.

\setlength{\belowcaptionskip}{-5pt}
\begin{table}[th!]
\noindent \centering{}%
\begin{tabular}{|c|c|c|}
\hline 
Dataset & MSE 2D & MSE 3D\tabularnewline
\hline 
\hline 
Online & $0.024m^{2}$ & $0.067m^{2}$\tabularnewline
\hline 
No SPBC & $0.026m^{2}$ & $0.071m^{2}$\tabularnewline
\hline 
No AS-ROI & $0.036m^{2}$ & $0.077m^{2}$\tabularnewline
\hline 
MAV 1 only & $0.061m^{2}$ & $0.140m^{2}$\tabularnewline
\hline 
MAV 2 only & $0.079m^{2}$ & $0.116m^{2}$\tabularnewline
\hline 
\end{tabular}\caption{\label{tab:Mean-Square-Error}Mean squared error (MSE) during the stationary person experiment.}
\end{table}
\setlength{\belowcaptionskip}{-5pt}

Figure~\ref{fig:Position-Error-over}, \ref{fig:Position-Error-over-1}
and Tab.~\ref{tab:Mean-Square-Error} present the results of the stationary person
experiment. Here, the MAV self-pose bias correction has
no significant overall effect on the person's 3D position estimate accuracy. There
are occasional situations in which not using the bias-corrected MAV self-pose
produces slightly better results than when using it (see, e.g., timestamp $112$ in Fig.~\ref{fig:Position-Error-over}).
However, this is outweighed by the risk of losing the person partially or completely if ROI is zoomed on the
wrong location (timestamp $98$ in Fig.~\ref{fig:Position-Error-over}). Disabling the active selection of ROI leads
to roughly comparable tracking accuracy to that of the online estimates, as long as the person is still
successfully detected. However, it has higher chances of detection failure, as is evident in the moving person experiment's results
(see Tab.~\ref{tab:Mean-Square-Error-1}). Consequently, it degrades the overall performance
(timestamp $72$ in Fig.~\ref{fig:Position-Error-over}) and results in a noticeably higher mean squared error.

As seen in Fig. \ref{fig:Position-Error-over-1}, cooperative perception significantly outperforms individual MAV estimates.

\subsection{\label{sub:Moving-Experiment}Moving person Experiment}

\begin{figure*}[t]
\noindent \begin{centering}
\includegraphics[width=1\textwidth]{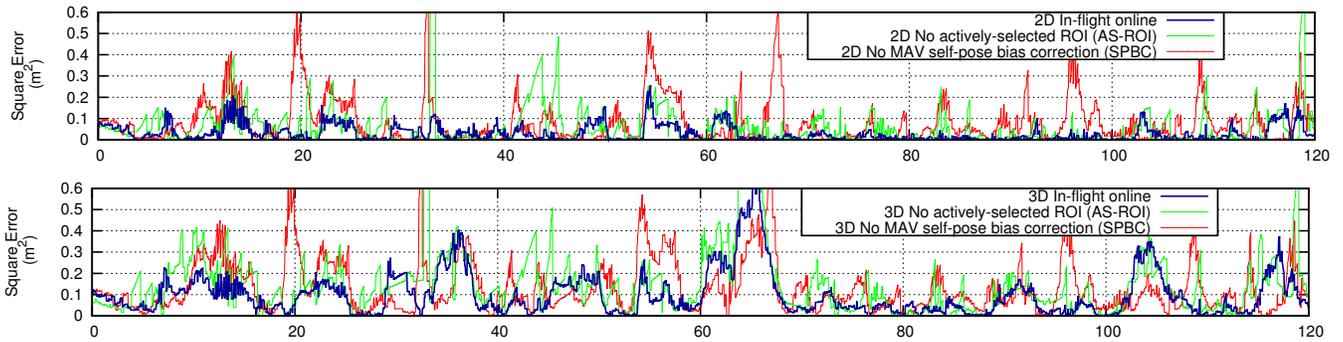}
\par\end{centering}

\noindent \centering{}\includegraphics[width=1\textwidth]{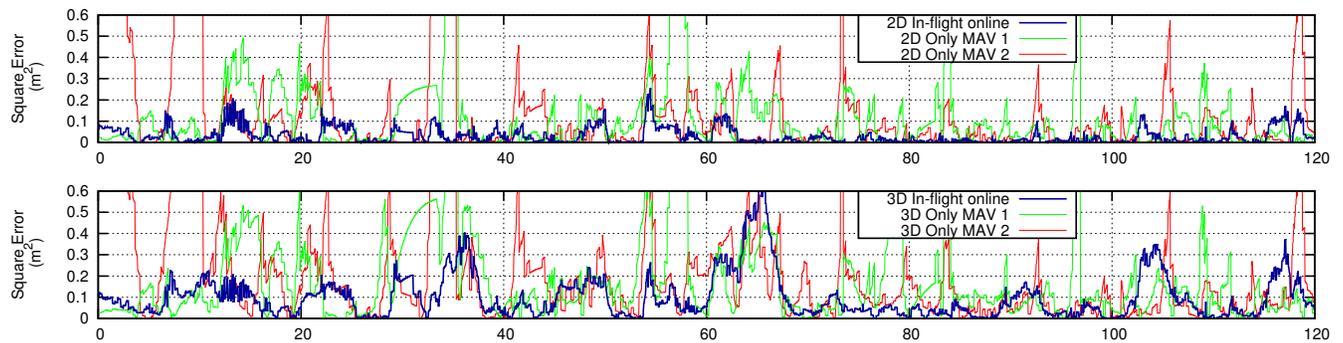}\caption{\label{fig:Position-Error-over-2}X Axis is time in seconds. Y axis represents 2D (only on the ground plane) or 3D position errors over time in the moving person experiment.}
\end{figure*}
\begin{figure*}[t]
\noindent \begin{centering}
\includegraphics[width=1\textwidth]{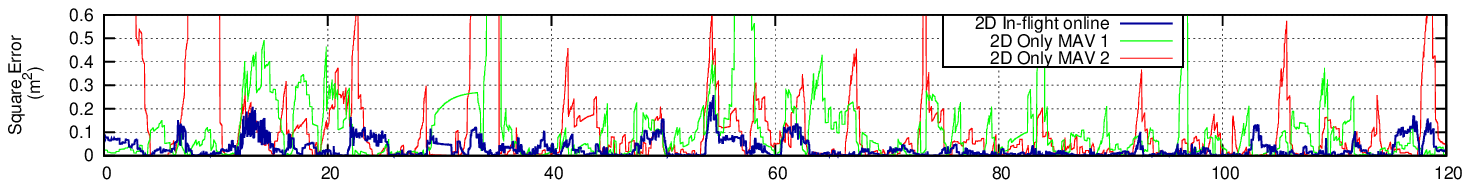}
\par\end{centering}

\noindent \begin{centering}
\includegraphics[width=1\textwidth]{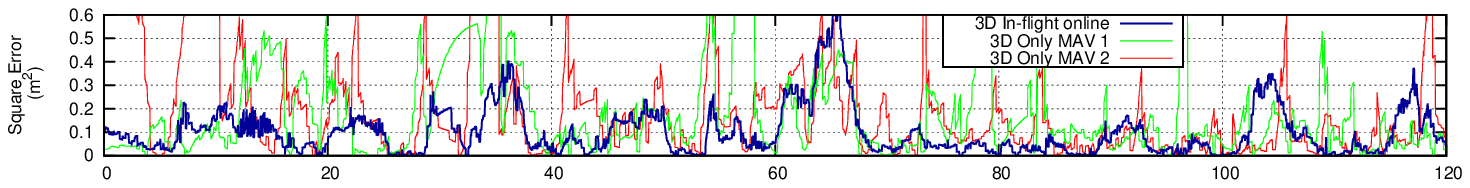}
\par\end{centering}

\noindent \centering{}\caption{\label{fig:Position-Error-over-1-1}X Axis is time in seconds. Y axis represents 2D (only on the ground plane) or 3D position errors over time in the moving person experiment.}
\end{figure*}
The formation constraints for the 2 MAVs are $D_{\textit{per}}$ = $8m$ and $H_{\textit{mav}}$ = $8m$.
Since we do not have the ground truth positions for the
person at any time, the person walks along a predefined trajectory, which is 
a square with $3m$ edge length aligned towards true north. The person
keeps walking the outline of this square for $120s$.

Considering this known trajectory constraint, we fit a horizontal,
north aligned reference square to each dataset (online and offline).
The error in a pose estimate is then calculated as the distance to
the closest point on this reference square. Fitting the reference
square is done in such a way that the mean squared error of all pose
estimates in each dataset is minimized.

\begin{table}[th!]
\centering{}%
\begin{tabular}{|c|c|c|c|c|}
\hline 
Dataset & MSE 2D & MSE 3D & NND MAV 1 & NND MAV 2 \tabularnewline
\hline 
\hline 
Online & $0.030m^{2}$ & $0.100m^{2}$ & $91.0\%$ & $97.5\%$\tabularnewline
\hline 
No SPBC & $0.081m^{2}$ & $0.132m^{2}$ & $91.0\%$ & $92.8\%$\tabularnewline
\hline 
No AS-ROI & $0.058m^{2}$ & $0.147m^{2}$ & $70.4\%$ & $63.9\%$\tabularnewline
\hline 
MAV 1 only& $0.138m^{2}$ & $0.218m^{2}$ & $89.5\%$ & $\circ$\tabularnewline
\hline 
MAV 2 only& $0.245m^{2}$ & $0.365m^{2}$ & $\circ$ & $97.7\%$\tabularnewline
\hline 
\end{tabular}\caption{\label{tab:Mean-Square-Error-1}Moving person experiment: MSE
and percentage of positive person detections over all frames on which detection was attempted.}
\end{table}

Fig, \ref{fig:Position-Error-over-2}, \ref{fig:Position-Error-over-1-1}
and table \ref{tab:Mean-Square-Error-1} show the results of the moving person 
experiment. Similar to the stationary experiment, the MAV self-pose bias correction
occasionally introduces additional errors, but overall it leads to better performance. 
The benefit of actively selecting ROI, the cornerstone of our MCDT method, is more significant  
in the dynamic situation of moving person and active formation following than in the stationary person experiment.
This is reflected in the mean squared errors calculated over the whole experiment.

Table \ref{tab:Mean-Square-Error-1} shows that active selection of ROI increases the chances of successful person detection by a large margin. 

Similar to the stationary person experiment, cooperative perception with 2 MAVs outperforms the single MAV case by factor of $\sim 2$ in the mean squared error of the person's 3D position estimate.

The video attached to this paper illustrates our work presented here and highlights a clip from the footage of the moving person experiment from a bird's eye view. A high resolution version of the attached video can be found here\footnote{\tiny \url{http://aircap.is.tuebingen.mpg.de/icra2018/video/}}. A more detailed video, showing the experimental setup, full stationary and moving person experiments, as well as the image streams from the MAV cameras overlaid with detections and EKF estimates can be found here\footnote{\tiny \url{http://aircap.is.tuebingen.mpg.de/icra2018/video_detailed/}}. Source code related to this work can be found on our project page\footnote{\tiny \url{http://aircap.is.tuebingen.mpg.de/}} and here\footnote{\tiny \url{http://aircap.is.tuebingen.mpg.de/icra2018/code/}}.

\section{Conclusions and Future Work}
\label{sec:conc} 

In this paper we presented a novel method for real-time, continuous and accurate DNN-based multi-robot cooperative detection and tracking.
Leveraging cooperation between robots in a team, our method is able to harness the power of deep convolutional neural network-based detectors for realtime applications.
Through real robot experiments involving only on-board computation and comparisons with baseline approaches, we demonstrated the effectiveness of our proposed method. We also showed the feasibility of real-time person detection and tracking with high accuracy from a team of MAVs which maintain a perception-driven formation. Additionally, we also performed noise quantification of the DNN-based detector that allowed us to use it within a Bayesian filter for person tracking.

The work in this paper paves the way for our future goal of performing full-body pose capture in outdoor unstructured scenarios. To this end, we are developing a perception-driven formation controller that not only attempts to minimize the joint 3D position uncertainty in realtime, but also considers the full-body pose reconstruction errors learned offline. Our future goals also include tracking animals and developing a heterogeneous team of MAVs with different sensing and maneuvering capabilities.

\bibliographystyle{IEEEtran}
\bibliography{paper}

\end{document}